\documentclass{article}

\usepackage{PRIMEarxiv}

\usepackage[utf8]{inputenc} 
\usepackage[T1]{fontenc}    
\usepackage{hyperref}       
\usepackage{url}            
\usepackage{booktabs}       
\usepackage{amsfonts}       
\usepackage{nicefrac}       
\usepackage{microtype}      
\usepackage{fancyhdr}       
\usepackage{graphicx}       
\graphicspath{{media/}}     
\usepackage{amsmath,amssymb,amsfonts}
\usepackage{multicol,multirow}
\usepackage{subcaption}

\title{GATGPT: A Pre-trained Large Language Model with Graph Attention Network for Spatiotemporal Imputation 
}

\author{
  Yakun Chen \\
  University of Technology Sydney \\
  Sydney\\
  \texttt{yakun.chen@student.uts.edu.au} \\
   \And
  Xianzhi Wang, Guandong Xu \\
  University of Technology Sydney \\
  Sydney\\
  \texttt{\{xianzhi.wang, guandong.xu\}@uts.edu.au} \\
}

\begin{document}
\maketitle

\begin{abstract}
The analysis of spatiotemporal data is increasingly utilized across diverse domains, including transportation, healthcare, and meteorology. In real-world settings, such data often contain missing elements due to issues like sensor malfunctions and data transmission errors. The objective of spatiotemporal imputation is to estimate these missing values by understanding the inherent spatial and temporal relationships in the observed multivariate time series. Traditionally, spatiotemporal imputation has relied on specific, intricate architectures designed for this purpose, which suffer from limited applicability and high computational complexity. In contrast, our approach integrates pre-trained large language models (LLMs) into spatiotemporal imputation, introducing a groundbreaking framework, GATGPT. This framework merges a graph attention mechanism with LLMs. We maintain most of the LLM parameters unchanged to leverage existing knowledge for learning temporal patterns, while fine-tuning the upper layers tailored to various applications. The graph attention component enhances the LLM's ability to understand spatial relationships. Through tests on three distinct real-world datasets, our innovative approach demonstrates comparable results to established deep learning benchmarks.
\end{abstract}


\section{Introduction}
The presence of multivariate time series data is extensively documented across a variety of sectors including economics, transportation, healthcare, and meteorology, as evidenced in several studies~\cite{bauer2015quiet,kaushik2020ai,yu2020stock,wu2019graph}. A range of statistical and machine learning techniques have been shown to perform effectively on complete datasets in several time series tasks, including forecasting~\cite{han2019review}, classification~\cite{ismail2019deep}, and anomaly detection~\cite{blazquez2021review}. However, it is often observed that multivariate time series data collected from real-world scenarios are prone to missing values due to various factors, such as sensor malfunctions and data transmission errors. These missing values can considerably affect the quality of the data, subsequently impacting the effectiveness of the aforementioned methods in their respective tasks.

Extensive research efforts have been dedicated to addressing the challenges in spatiotemporal imputation. A typical approach involves the development of a distinct framework for initially estimating missing values, followed by the application of the completed dataset in another sophisticated framework for subsequent operations like forecasting, classification, and anomaly detection. To fill in missing values, various statistical and machine learning techniques are applied. Popularly employed methods include autoregressive moving average (ARMA)~\cite{ansley1984estimation}, expectation-maximization algorithm (EM)~\cite{shumway1982approach}, and k-nearest neighbors (kNN)~\cite{oehmcke2016knn,hastie2009elements}. These models often depend on rigorous presumptions such as temporal uniformity and similarity between series, which may not be suitable for the complexity of real-world multivariate time series data, potentially leading to suboptimal performance.

Deep learning techniques have gained significant traction among researchers in the realm of spatiotemporal imputation. The utility of Recurrent Neural Networks (RNN) was initially recognized in this context~\cite{cao2018brits,che2018recurrent}. Subsequently, Generative Adversarial Networks (GAN) demonstrated impressive results, thanks to their robust generative properties~\cite{miao2021generative}. More recently, advancements have been made by integrating RNN-based methods with Graph Neural Networks (GNN) to enhance the capability of extracting spatial dependencies within these frameworks. Nevertheless, the current focus in this field is predominantly on developing intricate and specialized frameworks aimed at improving performance on specific public industrial datasets, often overlooking the aspects of generalizability and adaptability for varied datasets and downstream applications.

Additionally, the rise of pre-trained models in Natural Language Processing (NLP) has been noteworthy, with their applications extending into diverse areas such as Computer Vision (CV)~\cite{berrios2023towards}, Multi-modality~\cite{ye2023mplug}, and Recommendation Systems~\cite{zhang2023chatgpt}. These Large Language Models (LLMs) have demonstrated remarkable success due to their advanced representation learning abilities. Yet, there is a scarcity of research exploring using pre-trained LLMs in the domain of time series data analysis. Our research is focused on exploring the capacity of LLMs to act as potent representation learners, particularly in capturing temporal dependencies. Moreover, we recognize LLMs' exceptional few-shot learning capabilities, which positions them as highly suitable for time series contexts, where gathering extensive training data is often a formidable challenge.

In this paper, we aim to introduce the pre-trained Large Language Models (LLMs) to solve the spatiotemporal imputation problem. The main contributions of this paper are summarized as follows:

\begin{itemize}
    \item To the best of our knowledge, we are the first to introduce pre-trained LLMs to the spatiotemporal imputation. The integration of these models, with their pre-existing knowledge and architectural design, is instrumental in deciphering the intricacies of spatiotemporal dependencies.
    \item We have incorporated a graph attention module specifically designed to discern spatial dependencies. This module is devised to augment the LLMs in comprehending and assimilating the intrinsic characteristics of spatiotemporal data.
    \item Through our experimentation with various real-world datasets, we have demonstrated the effectiveness of our proposed framework. The results highlight its proficiency in spatiotemporal imputation, standing in comparison with multiple established baselines.
\end{itemize}

The rest of this paper is organized as follows. Section~\ref{sec:rel} reviews the related work on spatiotemporal imputation. Section~\ref{sec:method} presents the details of our proposed GATGPT framework, including the graph attention module and the pre-trained LLM block. Section~\ref{sec:exp} demonstrates our evaluation of the proposed framework performance on real-world datasets. Finally, Section~\ref{sec:con} draws our conclusions.

\section{Related Work}
\label{sec:rel}
\subsection{Spatiotemporal Imputation}

Addressing the spatiotemporal imputation challenge involves tackling both temporal and spatial dimensions. Initial research predominantly focused on using statistical and machine learning approaches for filling in missing data points~\cite{shumway1982approach,azur2011multiple,arumugam2018outlier}. Of late, deep learning techniques, particularly those based on recurrent neural networks (RNN), have gained prominence in spatiotemporal imputation due to their proficiency in capturing temporal patterns. Notable examples include GRU-D~\cite{che2018recurrent} and BRITS~\cite{cao2018brits}, which both utilize recurrent structures and analyze correlations among features.

To date, RNNs have been the cornerstone for many deep learning approaches in multivariate time series imputation, focusing on temporal dependency extraction. Additionally, Generative Adversarial Networks (GANs) have been explored for missing value prediction. GAIN~\cite{yoon2018gain}, for instance, uses a generator to impute data based on observed values and a discriminator to differentiate between observed and imputed sections. Another method, SSGAN~\cite{miao2021generative}, employs a semi-supervised classifier alongside a temporal reminder matrix to learn the data distribution for time series imputation.

GRIN~\cite{cini2021filling} combines graph neural networks with a recurrent structure, leveraging historical spatial patterns for imputation to simultaneously address spatial and temporal dependencies. However, a common limitation of these methods is their focus on developing specialized, task-specific frameworks. They often overlook the potential of incorporating pre-trained models, which could leverage inherent knowledge for improved inference capabilities in spatiotemporal imputation.

\subsection{Large Language Models for Time Series}
The efficacy of Large Language Models (LLMs) has been established across various disciplines, including computer vision, recommendation systems, and graph learning. Recent studies have increasingly explored the application of LLMs to multivariate time series tasks. PromptCast~\cite{xue2022promptcast} innovatively utilized prompt learning, a technique prevalent in natural language processing (NLP), for forecasting in time series. Gruver et al.~\cite{gruver2023large} used pre-trained LLMs as a zero-shot time series forester. More works focus on applying fine-tuning to LLMs to adjust the input and output suitable for time series~\cite{chang2023llm4ts,sun2023test,jin2023time}. In particular, Zhou et al.~\cite{zhou2023one} introduce comprehensive experiments in four different time series tasks: forecasting, classification, anomaly detection, and imputation. They commonly used GPT-2 as the backbone, froze attention and feedforward layers, and fine-tuned add and layer normalization layers. Lag-Llama~\cite{rasul2023lag} first applied Llama as the backbone of probabilistic time-series forecasting. All the above research gives us insights into considering pre-trained LLMs to deal with the spatiotemporal imputation.

\section{Methodology}
\label{sec:method}

\begin{figure}[!t]
\begin{center}
\includegraphics[width=0.6\textwidth]{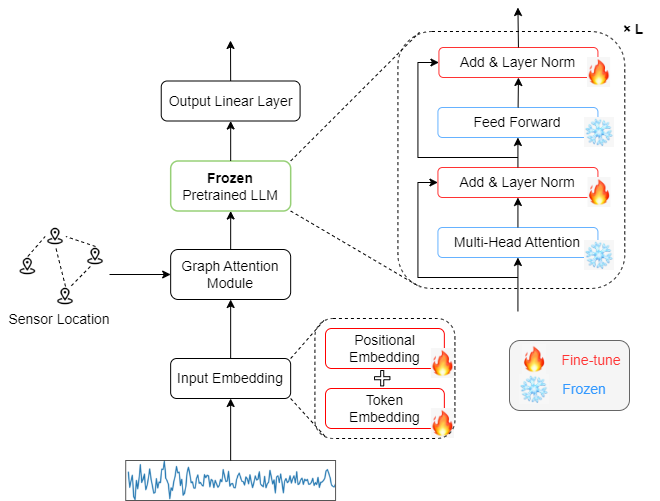}
\end{center}
\caption{The architecture of our proposed model.
} 
\label{fig:framework}
\end{figure}

Our proposed architecture, illustrated in Figure~\ref{fig:framework}, incorporates elements from pre-trained transformer models, specifically focusing on the GPT-2 model~\cite{radford2019language}, for spatiotemporal imputation. To effectively grasp spatial dependencies, we have integrated a graph attention module. This module functions by aggregating information from neighboring nodes into the node embeddings, which then serve as the input for the pre-trained Large Language Model (LLM). Furthermore, to bolster the model's capacity for discerning temporal relationships, we have implemented positional embedding into the original input. This addition is crucial for enabling the LLM to recognize the chronological sequence inherent in our spatiotemporal data.

\subsection{Input Embedding}
Our objective is to adapt pre-trained Large Language Models (LLMs) to a diverse range of tasks and modalities. This adaptation necessitates the redesign and retraining of the input embedding layer, which is intricately linked to the characteristics of the input data. Specifically, this layer is tailored to map time series data to the dimensional requirements of the chosen pre-trained model. In our approach, we employ two distinct types of embedding layers: token embedding and positional embedding. These layers are crucial for transforming the time series data into a format compatible with the pre-trained LLMs, facilitating effective learning and application across various tasks.

\begin{equation}
\begin{split}
    \textit{Emb}(x) &= \textit{TE}(x) + \textit{PE}(x) \\
    \textit{TE}(x) &= \textit{Conv1d}(x) \\
    \textit{PE}(pos, 2i) &= \sin\left(\frac{pos}{10000^{2i/d_{\text{model}}}}\right) \\
    \textit{PE}(pos, 2i+1) &= \cos\left(\frac{pos}{10000^{2i/d_{\text{model}}}}\right)
\end{split} 
\end{equation}
where \textit{TE} denotes the token embedding layer implemented by 1-dimensional convolution layer, \textit{PE} denotes the positional embedding layer, same with the vanilla transformer model. 

\subsection{Graph Attention Module}
Following the initial input embedding layer, we channel the embedded representation into the graph attention module. This module processes both the embedding and the adjacency matrix to yield an aggregated representation enriched with information from neighboring nodes. We construct the adjacency matrix by employing a thresholded Gaussian kernel similarity method, which is based on geographical distance information. This approach is in line with methodologies used in previous research, as detailed in the study by Cini et al.~\cite{cini2021filling}. The procedure operates as follows:

\begin{equation}
    \begin{split}
        \mathbf{H} &= \mathbf{X}\mathbf{W} \\
        e_{ij} &= \textit{LeakyReLU}(a^T[\mathbf{H}_i \Vert \mathbf{H}_j]) \\
        \alpha_{ij} &= \frac{\exp(e_{ij})}{\sum_{k \in \mathcal{N}(i)} \exp(e_{ik})} \\
        \textbf{H}'_i &= \textit{ELU}\left( \sum_{j \in \mathcal{N}(i)} \alpha_{ij} \textbf{H}_j \right) \\
        \textbf{H}''_i &= \Vert_{k=1}^K \textbf{H}'_{i,k}
    \end{split}
\end{equation}
where $\mathbf{W}$ is a learnable weight matrix, $\mathbf{X}$ is the input embedding representation. $a$ is learnable attention vector, $\alpha_{ij}$ is the attention weight after normalization, $\mathcal{N}(i)$ is the set of neighbor nodes for node $i$, $\textbf{H}'_i$ is updated node representation. In the multi-head graph attention layer with $k$ heads, we will calculate each node's representation and then concatenate them to generate the final aggregated representation. Especially, for the adjacency matrix, we applied DropEdge method to improve our model's generality.
\begin{equation}
    \mathbf{A}_{\text{dropped}} = \mathbf{A} \odot \mathbf{M}
\end{equation}
where $\mathbf{M}$ is a masked matrix with probability $p$ to drop the edges in the adjacency matrix $\mathbf{A}$.

\subsection{Frozen Pre-trained Blocks}

In our framework, we preserve the positional embedding layers and self-attention blocks from the pre-existing pre-trained models. Recognizing that the self-attention and feed-forward layers are repositories of a substantial portion of the knowledge accumulated in pre-trained language models, we choose to freeze the self-attention blocks. Meanwhile, we selectively fine-tune the addition and normalization layers to tailor the model to our specific requirements.

During the supervised fine-tuning phase, the model outputs representation embedding vectors for each node, essentially resembling a sequence of tokens. To address this, we integrate a linear layer which adjusts the final dimensional output to suit our needs.

In our empirical tests, we utilize GPT-2 as the foundational model. However, it's worth noting that our architecture is flexible and can incorporate other Large Language Models (LLMs), such as Llama, as alternatives. This versatility allows for the exploration of different model capabilities and adaptations in the context of spatiotemporal imputation.

\section{Experiment}
\label{sec:exp}
\subsection{Datasets}

For the evaluation of our GATGPT framework, we employed three distinct datasets, each relevant to the field of spatiotemporal imputation as outlined in previous research~\cite{cini2021filling}. The datasets include:

\begin{itemize}
    \item \textbf{AQI-36}: This dataset comprises hourly PM2.5 measurements recorded by 36 monitoring stations located in Beijing. The data span a period of 12 months, providing a comprehensive overview of air quality trends and variations over a yearly cycle.
    \item \textbf{PEMS-BAY}: This dataset features traffic speed data collected from 325 sensors strategically placed along highways in the San Francisco Bay Area. The data collection extends over a duration of 6 months, offering insights into traffic patterns and speed fluctuations in this region.
    \item \textbf{METR-LA}: It includes traffic speed data gathered by 207 sensors located on highways in Los Angeles County. The duration of data collection for this dataset is 4 months, capturing the traffic dynamics specific to the highways of Los Angeles.
\end{itemize}

Each of these datasets presents unique characteristics and challenges pertinent to spatiotemporal imputation, making them suitable for testing the efficacy and adaptability of the GATGPT framework.

\subsection{Baselines}
To assess the efficacy of our proposed GATGPT model, we conducted comparative analyses with both classic and cutting-edge models in the field of spatiotemporal imputation. The baseline models span a range of methodologies, including statistical (MEAN, DA, kNN, Linear), machine learning (MICE, VAR, KF), matrix factorization (TRMF, BATF), and deep learning (BRITS, GRIN, GP-VAE, rGAIN) techniques. Here is a brief overview of each baseline method:

\begin{itemize}
    \item \textbf{MEAN}: This method involves imputing missing values using the historical average value for each node.
    \item \textbf{DA}: Daily averages at corresponding time steps are used for imputation.
    \item \textbf{kNN}: This approach calculates and uses the average value of geographically proximate nodes for imputation.
    \item \textbf{KF}: A Kalman Filter is employed to impute time series data for each node.
    \item \textbf{MICE}~\cite{white2011multiple}: A multiple imputation method using chained equations.
    \item \textbf{VAR}: This is a vector autoregressive model used as a single step predictor.
    \item \textbf{TRMF}~\cite{yu2016temporal}: Temporal Regularized Matrix Factorization, a method for matrix factorization with temporal regularization.
    \item \textbf{BATF}~\cite{chen2019missing}: Bayesian Augmented Tensor Factorization, a model that integrates general forms of spatiotemporal domain knowledge.
    \item \textbf{BRITS}~\cite{cao2018brits}: A bidirectional RNN-based method for multivariate time series imputation.
    \item \textbf{GRIN}~\cite{cini2021filling}: This model uses a bidirectional GRU combined with a graph convolution network for multivariate time series imputation.
    \item \textbf{GP-VAE}~\cite{fortuin2020gp}: A probabilistic imputation method for time series that combines a Variational Autoencoder with Gaussian processes.
    \item \textbf{rGAIN}~\cite{yoon2018gain}: An approach that extends the GAIN model with a bidirectional recurrent encoder-decoder structure.

\end{itemize}

Each of these baseline models represents a different strategy or combination of techniques for spatiotemporal imputation, providing a comprehensive range for evaluating the performance of our GATGPT model.

\subsection{Evaluation Metrics}
In line with the methodology outlined in previous research~\cite{cini2021filling}, we utilize three well-established evaluation metrics to gauge the performance of our proposed GATGPT model in the context of spatiotemporal imputation. Our primary focus is on calculating the Mean Absolute Error (MAE) and Mean Squared Error (MSE) across the imputation window. Both MAE and MSE provide a quantitative measure of the absolute error between the values imputed by our model and the actual ground truth. These metrics are crucial for evaluating the accuracy and reliability of our model in accurately predicting missing values in spatiotemporal data.

\subsection{Experimental Settings}
\textbf{Datasets} Consistent with the dataset division strategy used in the study by Cini et al.~\cite{cini2021filling}, we have partitioned the training, validation, and test datasets for our evaluation. For the AQI-36 dataset, we have designated the data from March, June, September, and December as the test dataset. The validation dataset comprises the last 10\% of the data from February, May, August, and November. The remainder of the data is used for training purposes. Regarding the PEMS-BAY and METR-LA datasets, we have divided them chronologically into three segments: 70\% of the data is allocated for training, 10\% for validation, and the remaining 20\% serves as the test dataset. This structured approach ensures a comprehensive evaluation of our model across different time frames and varying conditions within each dataset.

\textbf{Imputation Strategies} For AQI-36 dataset, we adopt the same evaluation strategy as the previous work provided by~\cite{yi2016st}, which simulates the distribution of real missing data. For the traffic datasets PEMS-BAY and METR-LA, we designed a mask matrix to inject some missing values by hand, provided by~\cite{cini2021filling} for evaluation. There are two different missing patterns: (1) Point missing: randomly mask 25\% of observations. (2) Block missing: based on randomly masking 5\% of the observations, mask the remaining observations ranging from 1 to 4 hours for each sensor with probability of 0.15\%. It is important to note that in addition to these artificially injected missing values, all three datasets also contain inherent missing data, which amounts to 13.24\% in AQI-36, 0.02\% in PEMS-BAY, and 8.10\% in METR-LA. However, our evaluations focus solely on the missing portions artificially introduced into the test segments of each dataset, allowing for a controlled and precise assessment of our model’s imputation capabilities.

\subsection{Results}

\begin{table}[htbp]
  \centering
  \caption{Performance comparisons on three real-world datasets. The best metrics are highlighted in bold, while the subsequent best metrics are underscored.}
    \begin{tabular}{l|rr|rr|rr|rr|rr}
    \hline
\multicolumn{1}{c|}{\multirow{3}[2]{*}{Method}} & \multicolumn{2}{c|}{\multirow{2}[1]{*}{AQI-36}} & \multicolumn{4}{c|}{PEMS-BAY} & \multicolumn{4}{c}{METR-LA}\\
\cline{4-11}          &       &       & \multicolumn{2}{c|}{Point missing} & \multicolumn{2}{c|}{Block missing} & \multicolumn{2}{c|}{Point missing} & \multicolumn{2}{c}{Block missing}\\
\cline{2-11}          & \multicolumn{1}{l}{MAE} & \multicolumn{1}{l|}{MSE} & \multicolumn{1}{l}{MAE} & \multicolumn{1}{l|}{MSE} & \multicolumn{1}{l}{MAE} & \multicolumn{1}{l|}{MSE} & \multicolumn{1}{l}{MAE} & \multicolumn{1}{l|}{MSE} & \multicolumn{1}{l}{MAE} & \multicolumn{1}{l}{MSE}\\
    \hline
    Mean  & 53.48 & 4578.08 & 5.42  & 86.59 & 5.46  & 87.56 & 7.56  & 142.22 & 7.48  & 139.54 \\
    DA    & 50.51 & 4416.10 & 3.35  & 44.50 & 3.30  & 43.76 & 14.57 & 448.66 & 14.53 & 445.08 \\
    kNN   & 30.21 & 2892.31 & 4.30  & 49.80 & 4.30  & 49.90 & 7.88  & 129.29 & 7.79  & 124.61 \\
    \hline
    KF    & 54.09 & 4942.26 & 5.68  & 93.32 & 5.64  & 93.19 & 16.66 & 529.96 & 16.75 & 534.69 \\
    MICE  & 30.37 & 2594.06 & 3.09  & 31.43 & 2.94  & 28.28 & 4.42  & 55.07 & 4.22  & 51.07 \\
    VAR   & 15.64 & 833.46 & 1.30  & 6.52  & 2.09  & 16.06 & 2.69  & 21.10 & 3.11  & 28.00 \\
    TRMF  & 15.46 & 1379.05 & 1.85  & 10.03 & 1.95  & 11.21 & 2.86  & 20.39 & 2.96  & 22.65 \\
    BATF  & 15.21 & 662.87 & 2.05  & 14.90 & 2.05  & 14.48 & 3.58  & 36.05 & 3.56  & 35.39 \\
    \hline
    BRITS  & 14.50 & 622.36 & 1.47  & 7.94  & 1.70  & 10.50 & \underline{2.34}  & \underline{16.46} & \underline{2.34}  & \underline{17.00}\\
    GRIN  & 12.08 & 523.14 & \textbf{0.67}  & \textbf{1.55}  & \underline{1.14}  & \underline{6.60}  & \textbf{1.91}  & \textbf{10.41} & \textbf{2.03}  & \textbf{13.26} \\
    GP-VAE & 25.71 & 2589.53 & 3.41  & 38.95 & 2.86  & 26.80 & 6.57  & 127.26 & 6.55  & 122.33 \\
    rGAIN & 15.37 & 641.92 & 1.88  & 10.37 & 2.18  & 13.96 & 2.83  & 20.03 & 2.90  & 21.67\\
    \hline
    GATGPT & \textbf{10.28} & \textbf{341.26} & \underline{0.77} & \underline{1.77} & \textbf{0.77} & \textbf{1.77} & 2.64 & 21.32 & 2.66 & 21.52 \\
    \hline
    \end{tabular}%
  \label{tab:result}%
\end{table}%

In our study, we benchmarked the spatiotemporal imputation capabilities of the GATGPT model against various other baseline models. The comparative results, as depicted in Table~\ref{tab:result}, demonstrate that our proposed model attains a performance comparable to the GRIN model across all three datasets.

We observed that the statistical and classical machine learning methods tend to underperform in all datasets. This is largely attributed to their reliance on assumptions like stationarity or seasonality in time series, which are inadequate for capturing the more complex spatiotemporal correlations present in real-world datasets.

When it comes to deep learning methods, GATGPT exhibits superior performance on the AQI-36 dataset. In the case of the PEMS-BAY dataset, it ranks as the second-best in terms of performance, while for the METR-LA dataset, it holds the third-best position. This indicates that GATGPT is particularly effective in dealing with diverse types of spatiotemporal data, showcasing its robustness and versatility in various real-world scenarios.

\subsection{Analysis}
\begin{figure*}[t]
    \centering
    \begin{subfigure}[t]{0.48\textwidth}
        \centering
        \includegraphics[width=\textwidth]{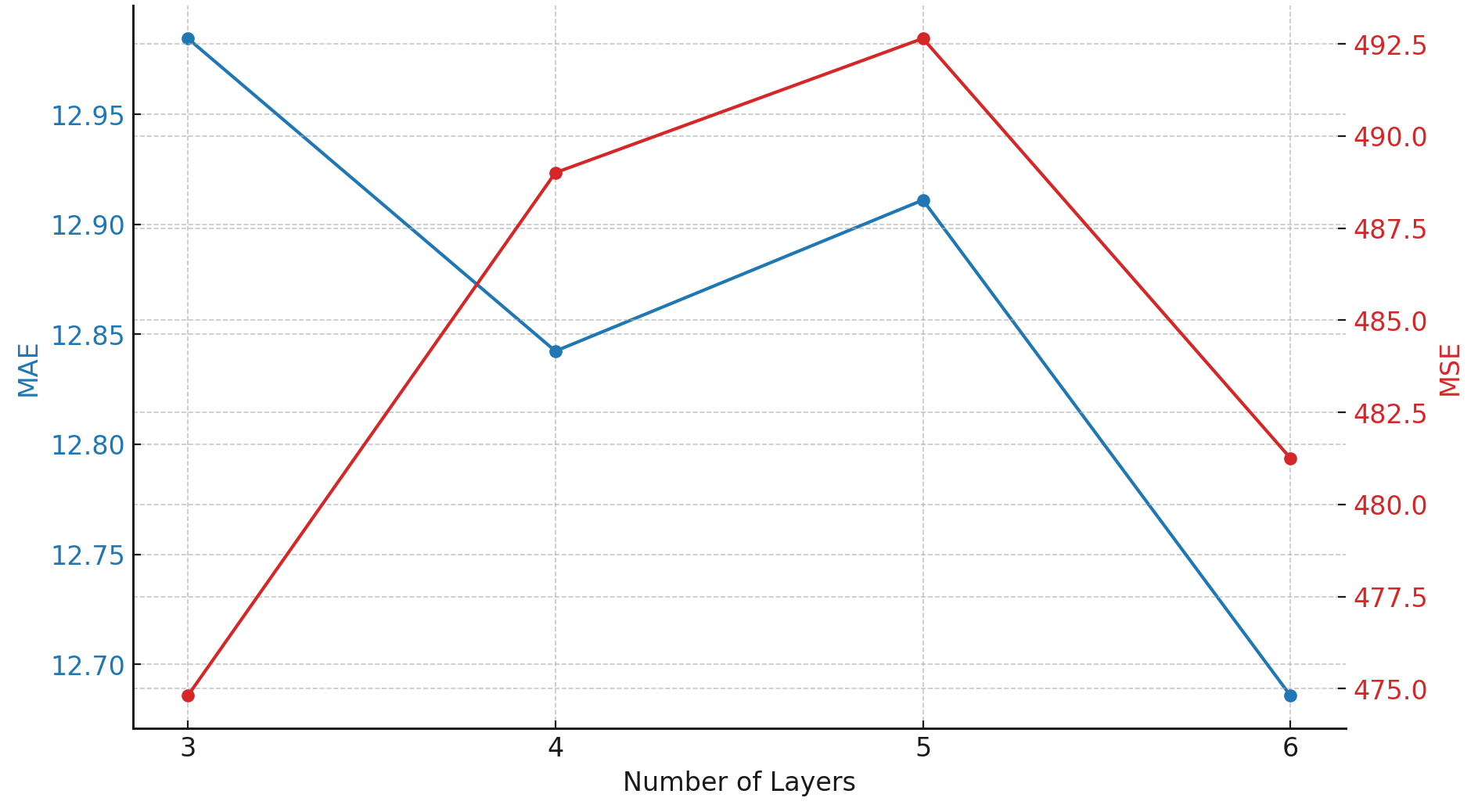}
        \caption{Performance on number of layers \{3, 4, 5, 6\}}
        \label{fig:layer}
    \end{subfigure}%
    ~ 
    \begin{subfigure}[t]{0.48\textwidth}
        \centering
        \includegraphics[width=\textwidth]{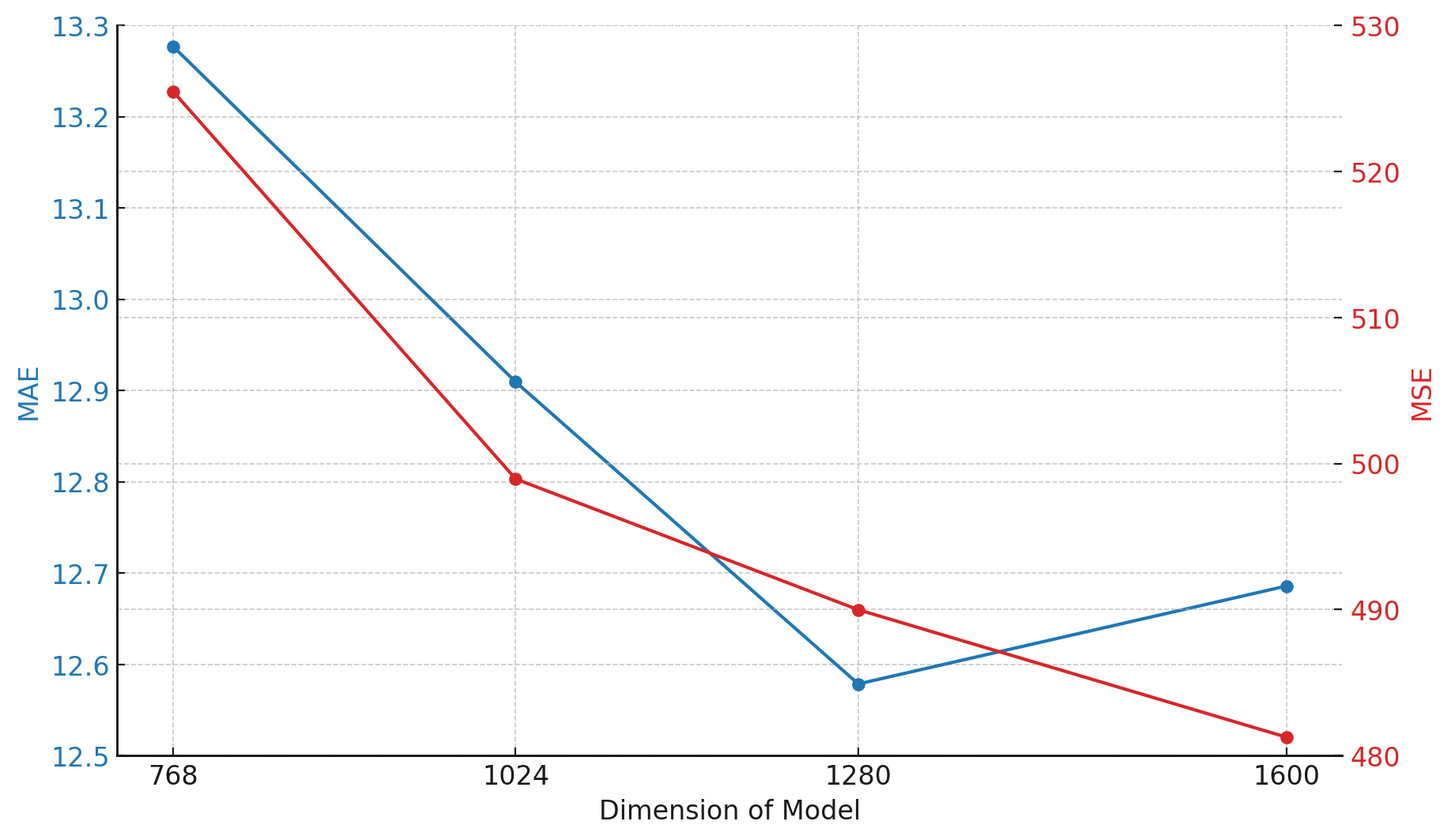}
        \caption{Performance on model dimension \{768, 1024, 1280, 1600\}}
        \label{fig:dmodel}
    \end{subfigure}
    \caption{The analysis for two key parameters with AQI-36 datasets.}
    \label{fig:para}
\end{figure*}

In this section of our study, we delve into sensitivity analyses focusing on key parameters of the pre-trained Large Language Model (LLM). Due to space constraints, Figure~\ref{fig:para} presents only a subset of our representative findings. However, it's worth noting that our observations from other datasets yield conclusions that are in line with these results.

Our discussion centers on two primary parameters and their impact on the model's performance: the number of layers in the LLM and the dimensionality of the pre-trained model. These parameters are crucial as they directly influence the model's ability to capture and process complex spatiotemporal patterns in the data. By adjusting and evaluating the effects of these parameters, we aim to gain deeper insights into the optimal configuration of the LLM for spatiotemporal imputation tasks. 

Figure~\ref{fig:para} shows the results of the parameter analysis on the AQI-36 dataset. Figure~\ref{fig:layer} showing MAE and MSE with model size of 1600 across different number of layers reveals an interesting pattern. Both MAE and MSE do not follow a straightforward linear trend with the increase in the number of layers. While there is an initial rise in errors as the layer count increases from 3 to 5, a notable decrease is observed at 6 layers, suggesting that more layers do not necessarily equate to lower errors. This non-linear relationship highlights the complexity of optimizing architectures, where the best performance may not be achieved by simply increasing the number of layers but rather by finding an optimal balance that minimizes errors. Figure~\ref{fig:dmodel} illustrates the relationship between model dimension and error metrics with the number of layers 6. It shows a clear trend: as model size increases from 768 to 1600, corresponding to GPT-2 Small, Medium, Large, and Extra Large, both MAE and MSE decrease, with the most notable improvement observed between 768 and 1024. However, the rate of improvement diminishes for larger dimensions, suggesting a point of diminishing returns. This indicates that while larger model sizes can enhance model accuracy, the benefits become less significant beyond a specific size, highlighting the need to balance computational efficiency with performance gains.

\section{Conclusion}
\label{sec:con}

In this paper, we present GATGPT, an innovative framework that combines graph attention networks and pre-trained Large Language Models (LLMs) for the purpose of spatiotemporal imputation. This framework is specifically designed to harness the capabilities of graph-based architectures to discern spatial relationships among various nodes. Subsequently, we align the pre-trained LLMs with the distinct characteristics of time series data, directing our focus toward spatiotemporal imputation tasks.

Our empirical investigations across three diverse real-world datasets from different sectors demonstrate the robust performance of the GATGPT framework. This success underscores the utility of pre-learned knowledge inherent in LLMs in deciphering complex spatiotemporal dependencies within datasets. The findings also highlight the potential of LLMs as a viable solution in real-world applications, particularly in scenarios characterized by few-shot learning and limited data availability. Our work thus positions LLMs as a powerful tool in the realm of spatiotemporal data analysis and imputation.


\bibliographystyle{unsrt}  
\bibliography{references}

\end{document}